\documentclass[preprint,12pt]{elsarticle}
\usepackage[T1]{fontenc}
\usepackage[utf8]{inputenc}
\usepackage{amsmath,amssymb,booktabs,tabularx}
\usepackage{xcolor}
\usepackage{float}
\usepackage{graphicx}
\usepackage{subcaption}
\usepackage{algorithm}
\usepackage{algpseudocode}
\usepackage{placeins}
\usepackage[hidelinks]{hyperref}
\journal{Robotics and Autonomous Systems}
\newcommand{\oasis}{\textsc{OASIS}}

\biboptions{numbers}
\begin{document}
\begin{frontmatter}
\title{OASIS: Optimization-Aware Subgoal Inference with Safety Constraints for Autonomous Mobile Robot Navigation}
\author[aff1]{Arian Hajizadeh}
\author[aff1]{Javad Pourrafi Arabani}
\author[aff1]{Negar Shenavar}
\author[aff1]{Sana Bashiri}
\author[aff1]{Seyed Majid Esmailzadeh\corref{cor1}}
\ead{smailzadeh@iust.ac.ir}

\cortext[cor1]{Corresponding author at: School of Electrical Engineering, Iran University of Science and Technology, Tehran, Iran}
\address[aff1]{Electrical Engineering, Iran University of Science and Technology, Tehran, Iran}
\begin{abstract}
Navigation in dynamic environments requires mobile robots
to make progress toward their goals while respecting motion
limits and avoiding surrounding obstacles. Learned subgoal
policies can guide model predictive control beyond its local
prediction horizon, but a geometrically suitable subgoal
does not necessarily lead to a successfully solved control
problem. This paper proposes OASIS, a hierarchical navigation
framework that incorporates recent optimization outcomes into
reinforcement learning-based subgoal generation. A high-level
policy trained using Soft Actor--Critic generates local
subgoals, while a nonlinear model predictive controller
tracks them subject to motion limits, geometric
collision-avoidance constraints, and embedded high-order
control barrier function constraints. The outcomes of
controller optimizations over each subgoal interval are
summarized and provided to the policy at its next decision,
without requiring additional optimization calls.
The framework is evaluated in five navigation scenarios
with static and moving obstacles, including dense
interactions and sensing uncertainty. Comparisons with
hierarchical and direct-goal control variants show that
OASIS achieves the highest aggregate navigation success.
Matched comparisons across three training seeds further
show a consistent success advantage over the otherwise
identical hierarchy without optimization-status input.
The selected-policy comparison also shows fewer primary
optimization failures with similar mean controller time.
\end{abstract}
\end{frontmatter}

\section{Introduction}

Safe navigation in dynamic environments remains a challenging
problem for autonomous mobile robots, requiring them to reach their goals while considering physical constraints and
adapting to the motion of surrounding obstacles. Despite extensive
progress in classical planning and control, uncertainty in
obstacle motion continues to challenge reliable navigation
\cite{shen2026motion}.

Model Predictive Control (MPC) has become an attractive approach to mobile robot motion planning and control because it can explicitly account for system constraints while optimizing future motion. Early applications focused on tracking predefined references. Ollero and Amidi demonstrated predictive path tracking using MPC \cite{Ollero1991PredictivePT}, while Klančar and Škrjanc developed a real-time trajectory-tracking controller with velocity and acceleration constraints \cite{Klanar2007TrackingerrorMP}. Bai et al. later compared different MPC-based tracking formulations, finding that linear formulations required less computation, while the nonlinear formulation handled demanding reference motions and positioning errors more effectively \cite{Bai2019ReviewAC}.

Recent approaches have addressed interactions with humans and incomplete environmental observations. Samavi et al. proposed SICNav for crowded environments, combining robot motion planning and human-motion prediction within a bilevel nonlinear MPC problem \cite{Samavi2025SICNavDiffusionSA}. Firoozi et al. developed OA-MPC for environments where occlusions may hide moving agents. Their method incorporates the forward reachable sets of potentially unseen agents into planning and introduces a terminal stopping constraint to establish recursive feasibility under the stated assumptions \cite{Firoozi2022OAMPCOM}. These approaches demonstrate how interaction modeling and occlusion awareness can inform predictive navigation.

Alongside these developments, researchers have investigated how safety requirements can be expressed within predictive control. While geometric constraints restrict the robot’s predicted positions relative to obstacles, control barrier functions (CBFs) impose conditions on the evolution of a safety function. Under appropriate assumptions, satisfying these conditions establishes forward invariance of the associated safe set. Ames et al. developed a framework combining CBF-based safety conditions and control Lyapunov function objectives within optimization-based controllers \cite{Ames2016ControlBF}. Zeng et al. incorporated discrete-time CBF constraints into MPC, combining predictive optimization and barrier-based safety conditions within a single formulation and analyzing its feasibility and stability \cite{Zeng2020SafetyCriticalMP}. Thirugnanam et al. further developed a discrete-time CBF formulation for collision avoidance between polytopic shapes, demonstrating obstacle avoidance with shorter prediction horizons than the distance-constrained formulations considered in their study \cite{Thirugnanam2021SafetyCriticalCA}.

For systems in which control inputs affect a safety function through higher-order derivatives, Xiao and Belta introduced high-order control barrier functions (HOCBFs). This extension is relevant to acceleration-controlled mobile robots, where control inputs influence position-based collision-avoidance constraints indirectly \cite{Xiao2021HighOrderCB}. More recently, Liu, Mao, and Belta combined discrete-time HOCBFs with an iterative MPC framework, constructing barrier constraints from grid-map-derived convex polytopes for navigation among dynamic obstacles \cite{Liu2024SafetyCriticalPA}. Other studies have addressed how barrier functions are constructed and adjusted during navigation. Salimi Lafmejani et al. incorporated barrier functions learned from LiDAR and odometry measurements into decentralized NMPC for navigation in unknown environments \cite{Lafmejani2022NMPCLBFNM}. Nguyen Minh et al. integrated relaxed CBF decay conditions and control Lyapunov function constraints into NMPC \cite{Minh2022EnhancingMR}. 

However, satisfying barrier conditions together with actuation and other system constraints can restrict the feasible solutions available to the controller. Zeng et al. investigated NMPC formulations with relaxed CBF conditions to address the relationship between feasibility and safety \cite{Zeng2021EnhancingFA}. Vangasse et al. compared MPC formulations incorporating ORCA and CBF constraints for distributed multi-robot path following through simulations and Crazyflie experiments. In their more demanding experimental scenario, frequent infeasible optimization problems limited the addition of further agents, while collision-avoidance behavior made returning to the reference path difficult \cite{Vangasse2024SafeNO}. These results highlight the need to consider not only the safety conditions imposed on a controller, but also optimization feasibility and the ability to make progress on the navigation task.  

Learning-based methods offer another way to improve navigation \cite{Wang2021ASO, Li2025HowTE}. Learning-based navigation methods can be organized according
to their role within the navigation pipeline: replacing the
entire pipeline, replacing individual subsystems, or improving
existing components~\cite{xiao2022motion}. End-to-end methods
learn motion commands directly from observations, reducing
reliance on separately designed planning and control modules. Tai et al. demonstrated an end-to-end approach to mapless navigation by training a reinforcement learning policy to generate linear and angular velocity commands directly from sparse laser measurements and the relative target position \cite{Tai2017VirtualtorealDR}.
However, a learned policy does not, by itself, ensure
satisfaction of motion and collision-avoidance constraints,
particularly outside its training conditions
\cite{xiao2022motion}. To address this limitation, hybrid approaches combine learned navigation policies with model-based safety mechanisms. Dawood et al. proposed an MPC-based safety shield that modifies the velocity commands generated by a reinforcement learning policy \cite{Dawood2024ADS}. Other approaches use learning to
adapt the parameters of a classical planner while retaining
its underlying structure~\cite{xiaoappl}. Their behavior
therefore remains dependent on the capabilities and
parameterization of that planner. 

Another form of hybrid navigation assigns subgoal generation to a learned high-level policy and motion execution to a model-based controller \cite{tolani2021visual,greatwood2019reinforcement}. MPC is well suited to this structure because it can track the selected references while explicitly accounting for system constraints. \cite{nascimento2018nonholonomic}. Brito et al. demonstrated this approach through an interaction-aware policy that recommends subgoals to an MPC planner \cite{brito2021}. This division allows learned guidance to complement local predictive control, while raising the question of how subgoal selection should account for the controller’s execution capabilities.

Subsequent studies have considered the compatibility between learned references and the downstream controller. Wen et al. restricted reference selection using a convex obstacle-free region and the robot’s kinematic limits, before tracking the resulting trajectory with MPC \cite{Wen2024CollisionFreeRN}. Huajian Liu et al. combined learned local-goal recommendations with spatio-temporal path search and MPC-based trajectory refinement, using incremental action masking to exclude goals that are unreachable or lead to infeasible optimization problems \cite{Liu2025HierarchicalLM}. These approaches account for execution constraints through reference-space construction or candidate evaluation.

\begin{figure}[t]
    \centering
    \includegraphics[width=0.65\linewidth]{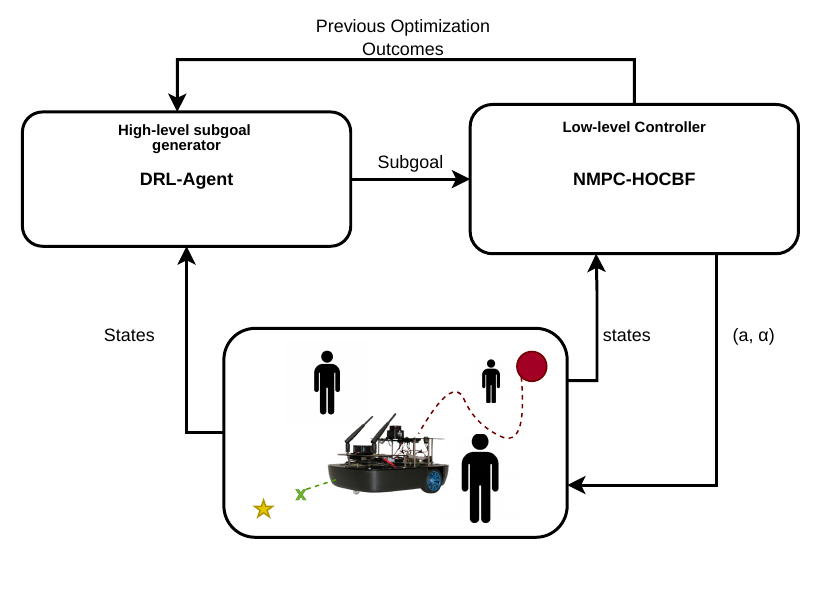}
    \caption{Proposed architecture. A high-level policy generates
subgoals from navigation observations and previous
optimization outcomes. A low-level predictive controller
tracks these subgoals subject to motion, collision-avoidance,
and HOCBF constraints, and returns optimization feedback
for the next policy decision.}
    \label{fig:oasisarch}
\end{figure}

These approaches motivate examining whether outcomes from previous controller optimizations can also inform subsequent subgoal selection. We propose \oasis{}, a hierarchical navigation framework that incorporates recent controller optimization outcomes into the high-level policy observation. As illustrated in
Fig.~\ref{fig:oasisarch}, the policy generates local
subgoals, while a nonlinear model predictive controller
(NMPC) computes control inputs subject to motion limits, and embedded HOCBF constraints. An aggregated
indicator of optimization outcomes during the preceding
subgoal interval is supplied to the policy at its next
decision. This feedback is obtained from existing controller
calls and requires neither additional optimization calls
nor an increase in the size of the optimization problem.
We investigate whether this explicit feedback improves
navigation when the controller and training reward are
held fixed.

The main contributions of this work are summarized as follows:
\begin{enumerate}
    \item An optimization-feedback interface that augments
    the subgoal policy observation with an aggregated
    indicator of controller optimization outcomes over
    the preceding subgoal interval.

    \item The integration of this interface into a
    hierarchical navigation framework combining learned
    subgoal generation, NMPC,
    and embedded HOCBF constraints for safety-constrained
    subgoal tracking.

    \item An empirical evaluation of optimization-status
    feedback through matched experiments across independent
    training runs, assessing its effect on navigation
    performance, solver reliability, and computational
    demand.
\end{enumerate}

\section{Preliminaries}
\subsection{System Dynamics}
\label{sec:sysdy}

To represent a vehicle's behavior, its continuous-time dynamics can be generalized as an affine nonlinear system:
\begin{equation}
    \dot{x} = f(x) + g(x)u
    \label{eq:statespace}
\end{equation}
where $x \in \mathbb{R}^n$ represents the state vector, $u \in \mathbb{R}^m$ is the control input, and the functions $f(x)$ and $g(x)$ characterize the nominal system dynamics. 

In this work, we consider a planar acceleration-controlled unicycle model. The state vector is defined by the 2D Cartesian position, heading angle, linear velocity, and angular velocity: $x = [p_x, p_y, \theta, v, \omega]^\top$. The system is controlled by a linear acceleration $a$ and an angular acceleration $\alpha$, forming the control input $u = [a, \alpha]^\top$. Its continuous dynamics are expressed as:
\begin{equation}
    \dot{x} = [v \cos\theta, v \sin\theta, \omega, a, \alpha]^\top
    \label{eq:dynamics}
\end{equation}

This model provides a highly convenient mathematical abstraction when computing Lie derivatives for forward invariance or formulating control barrier functions.

\subsection{Soft Actor-Critic}

Soft Actor--Critic (SAC) \cite{haarnoja2018soft} is an off-policy reinforcement
learning algorithm that learns a stochastic policy by
maximizing an entropy-regularized return:
\begin{equation}
\pi^\star =
\arg\max_{\pi}
\mathbb{E}_{\pi}
\left[
\sum_{t=0}^{\infty}\gamma^t
\left(
r(s_t,a_t)
+\beta\mathcal{H}\bigl(\pi(\cdot\mid s_t)\bigr)
\right)
\right],
\end{equation}
where $\gamma\in[0,1)$ is the discount factor and
$\beta>0$ controls the balance between reward and
policy entropy. The entropy term encourages exploration.

SAC uses an actor to represent the policy and two
critics to estimate action values. Training uses
transitions sampled from a replay buffer, together
with target critics to construct the learning targets.
The observation, action representation, and training
objectives used in OASIS are specified in
Section~\ref{sec:rl_subgoals}.

\subsection{High-order Control Barrier Function}
To mathematically guarantee safety, we define a Control Barrier Function, denoted as $h(x)$, which maps the system state to a real value such that $h(x) \ge 0$ represents a safe state .

The relative degree of a CBF with respect to the system dynamics is defined as the smallest integer $m$ for which the control input $u$ explicitly appears in the $m$-th time derivative of $h(x)$, meaning $L_g L_f^{m-1} h(x) \neq 0$.

For systems where the relative degree is greater than one, standard CBFs cannot be directly applied, necessitating the use of a HOCBF. The HOCBF formulation constructs a series of derived functions:
\begin{equation}
    \psi_i(x,t) = \dot{\psi}_{i-1}(x,t) + \alpha_i(\psi_{i-1}(x,t)), \quad i \in \{1, \dots, m\}
\end{equation}
where the initial condition is $\psi_0(x,t) = h(x,t)$, and each $\alpha_i$ is an extended class $\mathcal{K}_\infty$ function. 

For a relative-degree-$m$ constraint, define the
associated sets
\[
\mathcal{C}_i(t)
=
\{x:\psi_{i-1}(x,t)\geq 0\},
\qquad i=1,\ldots,m.
\]
The HOCBF condition requires
\begin{equation}
\sup_{u\in\mathcal{U}}\psi_m(x,u,t)\geq 0,
\qquad
x\in\bigcap_{i=1}^{m}\mathcal{C}_i(t),
\label{eq:hocbf_admissibility}
\end{equation}
where the derivatives in the recursive definition
are total derivatives along the system trajectories.
The controller enforces this condition by selecting
an admissible input satisfying $\psi_m(x,u,t)\geq 0$.

Specifically, for obstacle avoidance tasks, as an illustrative example, consider a point robot
avoiding a stationary circular obstacle. $h(x) = \frac{1}{2}[(x - x_o)^2 + (y - y_o)^2 - r_o^2]$, where $(x_o, y_o)$ denotes the position of the obstacle and $r_o$ represents its radius.  Consequently, by identifying class $\mathcal{K}$ functions $\alpha_1(\cdot)$ and $\alpha_2(\cdot)$, the HOCBF constraint for collision avoidance is simplified to:
\begin{equation}
    L_f^2 h(x) + L_g L_f h(x)u + L_f(\alpha_1(\psi_0(x))) + \alpha_2(\psi_1(x)) \ge 0
\end{equation}

The extension to moving obstacles used in this work
is presented in Section~\ref{sec:hocbf}.

\subsection{Nonlinear Model Predictive Control}To implement safety-critical optimal control over a future time horizon, we formulate a Receding Horizon Control (RHC) problem utilizing Nonlinear Model Predictive Control. It minimizes a predefined performance cost while strictly enforcing system constraints over a prediction horizon $N \ge 2$. Unlike standard linear MPC, the NMPC formulation explicitly handles the nonlinearities of the system. It requires the continuous-time nonlinear differential equation governing the system to be replaced by a discrete-time nonlinear difference equation to reduce computational complexity. The general NMPC problem is expressed as:

\begin{equation}
\begin{aligned}
\min_{X,U}\quad&
\sum_{h=0}^{N-1}
\ell(x_{h|k},u_{h|k})
+V_N(x_{N|k})\\
\text{subject to}\quad&
x_{0|k}=x_k,\\
&x_{h+1|k}=F_{\Delta t}(x_{h|k},u_{h|k}),
&&h=0,\ldots,N-1,\\
&u_{h|k}\in\mathcal U,
&&h=0,\ldots,N-1,\\
&x_{h|k}\in\mathcal X,
&&h=1,\ldots,N.
\end{aligned}
\label{eq:mpc_formulation}
\end{equation}

Here, $k$ denotes the discrete control time index,
$h$ indexes the prediction horizon, and $F_{\Delta t}$
is the discrete-time transition map. The functions
$\ell(\cdot)$ and $V_N(\cdot)$ denote the stage and
terminal costs, respectively. The sets $\mathcal U$
and $\mathcal X$ represent the acceleration and
velocity bounds. The terminal cost penalizes the
predicted position error relative to the current
subgoal.

\subsection{Problem formulation}\label{sec:problem}
As shown in \eqref{eq:dynamics}, we consider a planar acceleration-controlled unicycle model.
A fourth-order Runge--Kutta map $F_{\Delta t}$ propagates the state at sampling interval $\Delta t$. Linear and angular speeds and accelerations obey specified bounds. A goal is reached when the robot enters a goal-tolerance region, subject to separately recorded collision and boundary conditions.

Simulated moving obstacles follow crossing or oscillatory trajectories, depending on the scenario, with additional motion perturbations and sensing noise. Additionally, obstacles may change their velocity or direction.


Physical clearance from obstacle $i$ is defined as
\begin{equation}
c_i=\|p-o_i\|-r_R-\rho_i,
\end{equation}
where $\rho_i$ is the obstacle radius.
Contact or overlap is classified as a collision
when $c_i\leq0$. A violation of the prescribed
safety margin occurs when $c_i<m$.
A safety-margin violation therefore does not
necessarily imply physical contact.

\section{OASIS methodology}\label{sec:method}

\begin{figure}[t]
    \centering
    \includegraphics[width=0.95\linewidth]{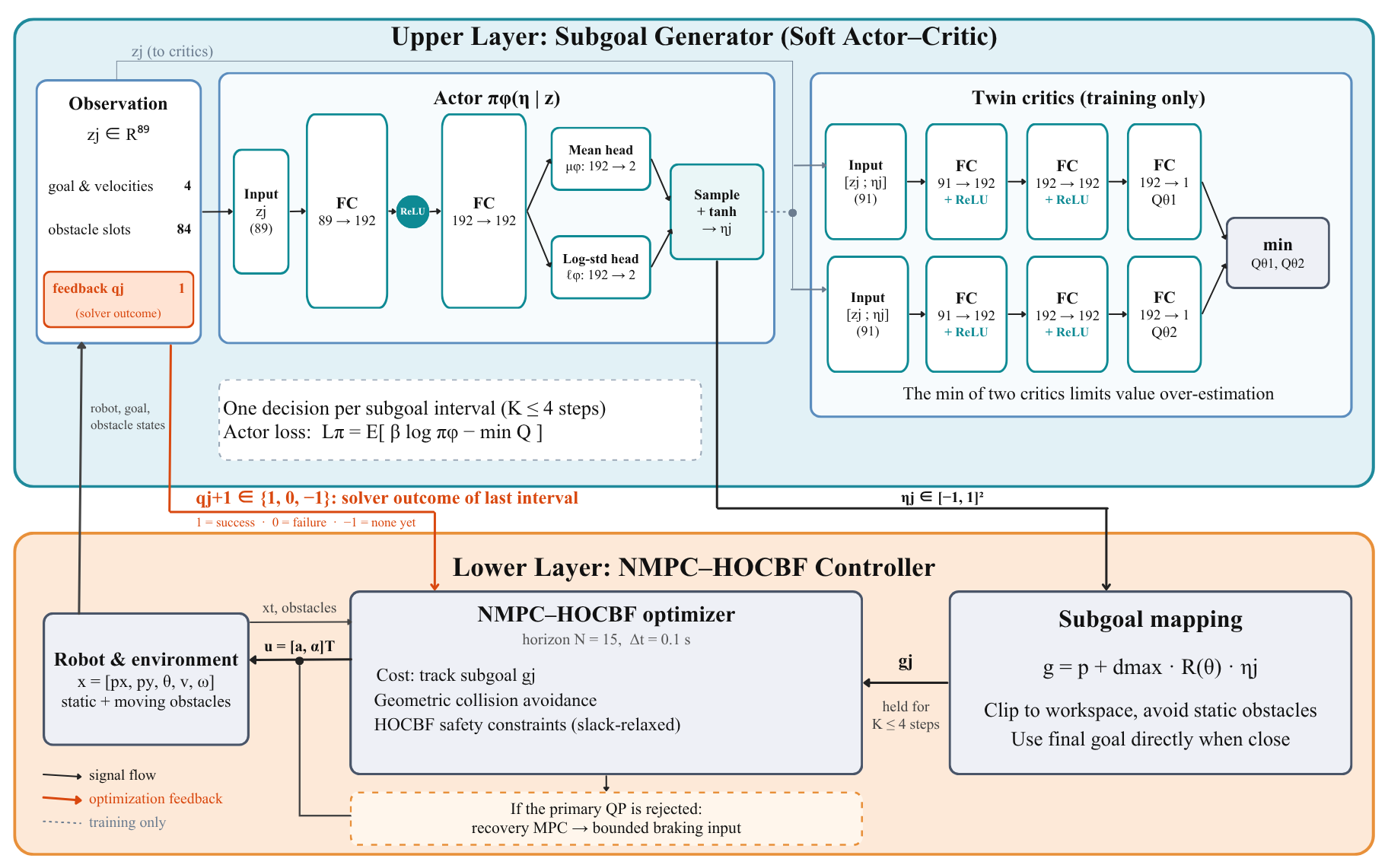} 
    \caption{Overall hierarchical architecture of OASIS. The upper layer uses a Soft Actor-Critic (SAC) framework to generate normalized subgoals $\eta_j$ based on environment observations and execution feedback $q_j$. The lower layer transforms $\eta_j$ into workspace subgoals $g_j$ and solves a Nonlinear Model Predictive Control problem with High-Order Control Barrier Functions (NMPC--HOCBF) to compute low-level control commands $\mathbf{u} = [a, \alpha]^T$ while returning solver outcomes $q_{j+1}$ to inform future subgoal generation.}
    \label{fig:oasissoftarch}
\end{figure}

In this paper, we propose a two-layer hierarchical architecture. The upper layer features a reinforcement learning-based subgoal generator that provides local sub-goals to the lower layer. In the lower layer, an MPC--HOCBF controller tracks
the generated subgoals subject to motion limits
and safety constraints. The complete mathematical formulation and methodology are detailed in the subsequent sections. The overall architecture of the proposed OASIS framework is illustrated in Fig.~\ref{fig:oasissoftarch}.

\subsection{Reinforcement learning for subgoal generation}
\label{sec:rl_subgoals}

The high-level policy generates local sub-goals that guide
the predictive controller toward the final navigation goal.
We train this policy using soft actor--critic (SAC).
The policy operates at a slower rate than the controller:
each subgoal is pursued for up to $K$ control steps before
a new subgoal is generated. We use $j$ to index high-level
decisions and $t$ to index low-level control steps.

\subsubsection{Policy observation}

At decision $j$, the policy receives an observation $z_j$
containing the goal location relative to the robot, the
robot's linear and angular velocities, obstacle features,
and the outcome of the previous optimization interval. Positions and velocities are expressed in the robot's
coordinate frame and normalized using fixed scales.

To accommodate different obstacle counts while retaining
a fixed observation dimension, we allocate $M$ obstacle
slots. Each occupied slot contains
\begin{equation}
z_j^{(i)}=
\left[
\tilde{r}_{i,x},
\tilde{r}_{i,y},
\tilde{c}_i,
\tilde{v}_{i,x}^{\mathrm{rel}},
\tilde{v}_{i,y}^{\mathrm{rel}},
d_i,
b_i
\right]^\top,
\end{equation}
where $\tilde{r}_i$ is the normalized relative obstacle
position, $\tilde{c}_i$ is the normalized physical clearance,
and $\tilde{v}_i^{\mathrm{rel}}$ is the normalized relative
velocity. The binary indicator $d_i$ distinguishes moving
obstacles from static obstacles, while $b_i$ indicates
whether the slot is occupied. Obstacles are ordered by
distance from the robot. Unoccupied slots are filled with
zeros and have $b_i=0$.

The complete observation is
\begin{equation}
z_j=
\left[
\tilde{g}_{j,x},
\tilde{g}_{j,y},
\tilde{v}_j,
\tilde{\omega}_j,
(z_j^{(1)})^\top,\ldots,(z_j^{(M)})^\top,
q_j
\right]^\top,
\label{eq:rl_observation}
\end{equation}
where $\tilde{g}_j$ denotes the normalized relative
position of the final goal and $q_j$ is the optimization
feedback. Its dimension is therefore $4+7M+1$.
The no-feedback variant omits $q_j$.

\subsubsection{Optimization feedback}

The policy receives a summary of the primary controller's
optimization outcomes from the preceding subgoal interval.
Let $L_j$ denote the number of controller calls executed
while pursuing subgoal $g_j$, and let $N_j^{\mathrm{fail}}$
denote the number of those calls that fail to return an
accepted solution. The feedback available at the next
policy decision is
\begin{equation}
q_{j+1} =
\begin{cases}
1,  & L_j > 0 \text{ and } N_j^{\mathrm{fail}} = 0,\\
0,  & L_j > 0 \text{ and } N_j^{\mathrm{fail}} > 0,\\
-1, & L_j = 0.
\end{cases}
\label{eq:optimization_feedback}
\end{equation}

At the beginning of each episode, $q_0=-1$ indicates
that no previous controller outcome is available.
An accepted solution must satisfy the solver-status
and numerical acceptance checks of the controller.
A failed primary solve remains counted as a failure
even when a subsequent recovery action is available.

At decision $j$, the policy generates its action from
the navigation observation and $q_j$. The outcomes
associated with this action are summarized in $q_{j+1}$
and become available at the following policy decision.
The signal is computed from existing controller calls
and requires no additional optimization.

We distinguish optimization feedback in the policy
observation from optimization-related reward penalties.
The observation communicates the previous interval's
outcome to the policy, whereas the reward penalties
influence policy learning.

\subsubsection{Action and subgoal mapping}

The policy produces a two-dimensional action
$\eta_j\in[-1,1]^2$, representing a local displacement
in the robot's coordinate frame. The action is mapped
to a requested subgoal according to
\begin{equation}
g_j^{\mathrm{req}}
=
p_{t_j}
+
d_{\max}\mathcal{R}(\theta_{t_j})
\frac{\eta_j}{\max(1,\|\eta_j\|_2)},
\label{eq:rl_subgoal_mapping}
\end{equation}
where $d_{\max}$ is the maximum subgoal distance and
\begin{equation}
\mathcal{R}(\theta)=
\begin{bmatrix}
\cos\theta & -\sin\theta\\
\sin\theta & \cos\theta
\end{bmatrix}
\end{equation}
rotates the displacement into world coordinates.

The requested subgoal is subsequently processed to respect
the configured workspace limits and avoid endpoints inside
inflated static obstacles. When the robot is sufficiently
close to the final goal and the connecting segment is clear
of static obstacles, the final goal is used directly.
The resulting subgoal $g_j$ is passed to the low-level
controller, which computes the physical control inputs.

\subsubsection{Reward}

The high-level reward combines the navigation rewards
accumulated during the subgoal interval with penalties
for optimization failures and emergency stops:
\begin{equation}
\begin{aligned}
R_j={}&
\sum_{\ell=0}^{L_j-1}
\Big[
w_p(d_t-d_{t+1})-w_t-w_u\|u_t\|_2^2\\
&\qquad
+w_g\mathbf{1}_{\mathrm{goal}}
-w_c\mathbf{1}_{\mathrm{collision\ or\ boundary}}
\Big]\\
&-w_f\frac{N_j^{\mathrm{fail}}}{\max(L_j,1)}
-w_s\frac{N_j^{\mathrm{stop}}}{\max(L_j,1)},
\end{aligned}
\label{eq:rl_reward}
\end{equation}
where $t=t_j+\ell$, $d_t$ is the distance to the final
goal, and $L_j\leq K$ is the number of executed control
steps. The counts $N_j^{\mathrm{fail}}$ and
$N_j^{\mathrm{stop}}$ denote primary optimization failures
and emergency stops during the interval.

The progress term rewards movement toward the goal.
The time and control penalties discourage unnecessarily
long episodes and excessive control effort, while the
terminal terms reward arrival and penalize collision
or boundary violation. We use $w_f=0.5$ and $w_s=1.0$.
No reward term explicitly depends on solver execution time.

The optimization-related reward terms and the feedback
observation serve different roles. The failure fraction
penalizes unsuccessful primary optimization calls, while
the stopping fraction penalizes emergency stops. Neither
fraction is directly included in the policy observation;
the observation contains the summary $q_j$ defined in
\eqref{eq:optimization_feedback}.

\subsubsection{Soft actor--critic training}

SAC learns a stochastic policy $\pi_\phi(\eta\mid z)$
and two action-value functions $Q_{\theta_1}$ and
$Q_{\theta_2}$. Its entropy-regularized objective is
\begin{equation}
\max_\phi\;
\mathbb{E}_{\pi_\phi}
\left[
\sum_{j=0}^{\infty}\gamma^j
\left(
\bar{R}_j+
\beta\mathcal{H}\bigl(\pi_\phi(\cdot\mid z_j)\bigr)
\right)
\right],
\label{eq:sac_objective}
\end{equation}
where $\gamma$ is the discount factor, $\beta>0$ is
the entropy temperature, and
$\bar{R}_j=c_R R_j$ is the scaled reward used for learning.

Transitions $(z_j,\eta_j,\bar{R}_j,z_{j+1},d_j)$
are stored in a replay buffer. Here, $d_j$ indicates
a terminal transition; time-limit truncations retain
bootstrapping. The target for the critic updates is
\begin{equation}
\begin{aligned}
y_j=\bar{R}_j+\gamma(1-d_j)
\Big[
&\min_{i\in\{1,2\}}
Q_{\bar{\theta}_i}(z_{j+1},\eta')\\
&-\beta\log\pi_\phi(\eta'\mid z_{j+1})
\Big],
\end{aligned}
\end{equation}
with $\eta'\sim\pi_\phi(\cdot\mid z_{j+1})$ and
target-network parameters $\bar{\theta}_i$.
Each critic minimizes
\begin{equation}
\mathcal{L}_{Q_i}
=
\mathbb{E}
\left[
\left(Q_{\theta_i}(z_j,\eta_j)-y_j\right)^2
\right].
\end{equation}
The actor minimizes
\begin{equation}
\mathcal{L}_{\pi}
=
\mathbb{E}
\left[
\beta\log\pi_\phi(\eta\mid z)
-
\min_{i\in\{1,2\}}Q_{\theta_i}(z,\eta)
\right],
\end{equation}
where $\eta\sim\pi_\phi(\cdot\mid z)$.
The entropy temperature is adapted during training,
and the target critics are updated by
\begin{equation}
\bar{\theta}_i
\leftarrow
\tau\theta_i+(1-\tau)\bar{\theta}_i.
\end{equation}

Training additionally uses heuristic demonstrations,
teacher actions during data collection, and an auxiliary
imitation loss, as detailed in
Section~\ref{sec:heuristic_guidance}

During evaluation, the policy parameters are fixed and
the action is computed deterministically as
\begin{equation}
\label{eq:deterministic_action}
\eta_j=\tanh\bigl(\mu_\phi(z_j)\bigr).
\end{equation}
No exploratory action sampling or heuristic action
blending is used during evaluation.

\subsubsection{Neural network architecture}
\label{sec:network_architecture}

The learning architecture consists of a stochastic actor,
two independently parameterized critics, and two target
critics. All networks are fully connected feedforward
networks. The actor generates subgoal actions, while the
critics estimate their expected entropy-regularized return.
Only the actor is required for action generation during
deployment.

\paragraph{Actor network}
The actor receives the observation $z_j\in\mathbb{R}^{n_z}$.
With $M=12$ obstacle slots, the observation dimension is
$n_z=89$ for OASIS and $n_z=88$ when optimization feedback
is omitted. The observation passes through two hidden
layers containing 192 units each:
\begin{align}
h_1 &= \operatorname{ReLU}(W_1z_j+b_1),\\
h_2 &= W_2h_1+b_2.
\end{align}
Two separate linear output heads produce the mean and
logarithmic standard deviation of a two-dimensional
Gaussian distribution:
\begin{align}
\mu_\phi(z_j) &= W_\mu h_2+b_\mu,\\
\ell_\phi(z_j) &=
\operatorname{clip}(W_\sigma h_2+b_\sigma,-5,2),\\
\sigma_\phi(z_j) &= \exp\bigl(\ell_\phi(z_j)\bigr).
\end{align}
The covariance is diagonal, with entries given by
$\sigma_\phi^2(z_j)$.

During stochastic action generation, the actor uses
the reparameterization
\begin{equation}
\eta_j=
\tanh\left(
\mu_\phi(z_j)+
\sigma_\phi(z_j)\odot\epsilon_j
\right),
\qquad
\epsilon_j\sim\mathcal{N}(0,I_2),
\end{equation}
where $\odot$ denotes element-wise multiplication.
The hyperbolic tangent bounds both action components
to $[-1,1]$. These components represent a local subgoal
displacement rather than direct robot control inputs.
The SAC log-probability calculation accounts for this
nonlinear squashing transformation.
During evaluation, the deterministic action is obtained
from the actor mean according to
Equation~\eqref{eq:deterministic_action}.

\paragraph{Critic networks}
Each critic receives the concatenation of the observation
and the two-dimensional action:
\begin{equation}
\xi_j=[z_j^\top,\eta_j^\top]^\top.
\end{equation}
Its input dimension is therefore 91 for OASIS and 90
for the no-feedback variants. Each critic contains
two hidden layers of 192 units, with ReLU activation
after both layers, followed by a scalar linear output:
\begin{align}
c_{i,1}
&=\operatorname{ReLU}(W_{i,1}\xi_j+b_{i,1}),\\
c_{i,2}
&=\operatorname{ReLU}(W_{i,2}c_{i,1}+b_{i,2}),\\
Q_{\theta_i}(z_j,\eta_j)
&=W_{i,3}c_{i,2}+b_{i,3},
\qquad i\in\{1,2\}.
\end{align}
The critics have separate parameters, and their minimum
estimate is used in the SAC updates to reduce value
overestimation. Each target critic has the same
architecture as its corresponding critic and is
updated through exponential averaging.

\paragraph{Training configuration}
The actor, critics, and entropy temperature are optimized
using Adam with learning rates of $3\times10^{-4}$.
Training uses mini-batches of 64 transitions sampled from
a replay buffer with capacity 120,000. The discount
factor is $\gamma=0.99$, the target-network update
coefficient is $\tau=0.005$, and the reward scaling
factor is $c_R=0.1$. The entropy temperature is learned
using a target entropy of $-2$.

The networks contain no recurrent state. Consequently,
optimization feedback provides an explicit summary of
the previous controller interval within the current
observation.

\subsubsection{Heuristic-guided policy training}
\label{sec:heuristic_guidance}

A heuristic subgoal generator provides initial guidance
during policy training. It evaluates candidate directions
around the direction of the final goal, favoring goal
progress while penalizing predicted proximity to obstacles.
The scoring also accounts for obstacle velocities to
discourage motion in front of crossing obstacles.
The highest-scoring candidate provides the teacher
action $\eta^H(z)$.

The heuristic supports learning in three ways. First,
demonstration trajectories and synthetic observation--action
pairs are used to initialize the actor through behavior
cloning. Second, heuristic actions are occasionally used
during online data collection. Third, an auxiliary
imitation loss encourages the actor to retain useful
behavior from the demonstrations while improving through
reinforcement learning. This use of demonstrations is
related to demonstration-assisted policy-learning
approaches~\cite{rajeswaran2018learning}.

At training episode $e$, the action used for data
collection is
\begin{equation}
\eta_j =
\begin{cases}
\eta^H(z_j), & \text{with probability } p_H(e),\\
\eta_j^\pi,\quad \eta_j^\pi\sim\pi_\phi(\cdot\mid z_j),
& \text{otherwise},
\end{cases}
\end{equation}
where
\begin{equation}
p_H(e)=
\max\left(
0.05,\,
0.25\left(1-\frac{e}{E_{\mathrm{sched}}}\right)
\right).
\end{equation}
Here, $E_{\mathrm{sched}}$ is the episode count used
to define the guidance schedule. The matched
1,600-episode experiments use $E_{\mathrm{sched}}=5000$.

The actor objective includes the demonstration loss
\begin{equation}
\mathcal{L}_{\mathrm{actor}}
=
\mathcal{L}_{\pi}
+
\lambda_{\mathrm{BC}}(n)
\mathbb{E}_{(z,\eta^H)\sim\mathcal{D}_H}
\left[
\|\tanh(\mu_\phi(z))-\eta^H\|_2^2
\right],
\end{equation}
where $\mathcal{D}_H$ contains heuristic demonstration
pairs and $n$ denotes the number of learning updates.
The imitation weight follows
\begin{equation}
\lambda_{\mathrm{BC}}(n)
=
3\max\left(0.05,1-\frac{n}{5000}\right).
\end{equation}

Prior control knowledge can support policy learning
in robot navigation. Han et al.~\cite{han2025drmpc}
proposed DR-MPC, which initializes navigation with
MPC-based path tracking and learns to blend MPC
with reinforcement learning-based actions.
In OASIS, prior knowledge is introduced through
heuristic demonstrations, teacher actions during
data collection, and an auxiliary imitation loss.
During evaluation, the trained actor generates
subgoals without heuristic action blending,
while MPC remains responsible for constrained
motion control.




\subsection{Low Level}
The low-level controller tracks the subgoal generated
by the high-level policy. It computes control inputs
subject to robot motion limits, and HOCBF constraints.

\subsubsection{NMPC}
\label{sec:nmpc}

The low-level controller tracks the subgoal $g_j$ using the robot dynamics defined in Section~\ref{sec:sysdy}. The discrete prediction model is defined as
\begin{equation}
x_{k+1}=F_{\Delta t}(x_k,u_k),
\qquad x_0=x(t),
\label{eq:discrete_prediction}
\end{equation}
where $F_{\Delta t}$ denotes the discrete-time
transition map over the sampling interval $\Delta t$.
Over a horizon of $N$ steps, the controller minimizes
\begin{align}
J_{\mathrm{MPC}}
={}&
\sum_{k=0}^{N-1}
\left(
\|x_k-x_k^{\mathrm{ref}}\|_Q^2
+\|u_k\|_{R_u}^2
+\|u_k-u_{k-1}\|_{R_\Delta}^2
\right)
\nonumber\\
&+\|p_N-g_j\|_{Q_f}^2,
\label{eq:nmpc_cost}
\end{align}
where $\|z\|_W^2=z^\top Wz$, $p_k=[p_{x,k},p_{y,k}]^\top$,
and $u_{-1}$ is the previously applied input. The reference
state is
\[
x_k^{\mathrm{ref}}
=
[g_{j,x},g_{j,y},\theta_k^{\mathrm{ref}},
v_k^{\mathrm{ref}},0]^\top,
\]

subject to

\begin{align}
\nonumber u_{\min}&\leq u_k\leq u_{\max},
&& k=0,\ldots,N-1,\\
\nonumber v_{\min}&\leq v_k\leq v_{\max},
&& k=1,\ldots,N,\\
\nonumber \omega_{\min}&\leq\omega_k\leq\omega_{\max},
&& k=1,\ldots,N,
\end{align}

The reference heading points from the nominal predicted
position toward the subgoal, with the angular difference
wrapped to $[-\pi,\pi)$. The reference speed is
proportional to the remaining distance and capped
at $v_{\max}$:
\[
v_k^{\mathrm{ref}}
=
\min\!\left(v_{\max},
k_v\|g_j-\bar p_k\|\right),
\qquad k_v=1\,\mathrm{s}^{-1},
\]
where $\bar p_k$ denotes the nominal predicted position.
These references are computed from the nominal trajectory
and held fixed during the QP solve.

For obstacle $i$, let $o_i$, $v_i^o$, and $\rho_i$ denote
its observed position, velocity, and radius. The controller employs the local constant-velocity
forecast. Its predicted
position is

\begin{equation}
\hat{o}_{i,k}=o_i+k\Delta t\,v_i^o.
\end{equation}
Geometric obstacle avoidance is imposed through
\begin{equation}
\|p_k-\hat{o}_{i,k}\|^2
\geq
(r_R+\rho_i+\varepsilon)^2,
\qquad k=1,\ldots,N,
\label{eq:nmpc_geometric}
\end{equation}
where $r_R$ is the robot radius and $\varepsilon$ is the
geometric clearance offset.

\subsubsection{HOCBF}
\label{sec:hocbf}

The safety constraints use a buffered robot--obstacle
separation. Define
\begin{equation}
h_i=r_i^\top r_i-R_i^2,
\qquad
r_i=p-o_i,
\qquad
R_i=r_R+\rho_i+m,
\label{eq:barrier}
\end{equation}
where $m$ is the safety margin. The associated safe set is
\[
\mathcal{C}_i=\{x:h_i(x,t)\geq0\}.
\]

Let
\[
e(\theta)=
\begin{bmatrix}\cos\theta\\\sin\theta\end{bmatrix},
\qquad
e_\perp(\theta)=
\begin{bmatrix}-\sin\theta\\\cos\theta\end{bmatrix}.
\]
The robot's translational velocity and acceleration are
\begin{equation}
v_R=ve(\theta),
\qquad
a_R=ae(\theta)+v\omega e_\perp(\theta).
\end{equation}
Consequently,
\begin{equation}
\dot{h}_i=2r_i^\top\bigl(ve(\theta)-v_i^o\bigr).
\end{equation}

Under the obstacle-acceleration bound
$\|a_i^o\|\leq\bar a_i^o$, the second derivative satisfies
$\ddot h_i\geq\underline{\ddot h}_i$, where
\begin{align}
\underline{\ddot h}_i
={}&2\|ve(\theta)-v_i^o\|^2
+2r_i^\top
\left(ae(\theta)+v\omega e_\perp(\theta)\right)
\nonumber\\
&-2\|r_i\|\bar a_i^o.
\label{eq:barrier_acceleration_bound}
\end{align}
For static obstacles, $v_i^o=0$ and $\bar a_i^o=0$.

With positive gains $\gamma_1$ and $\gamma_2$, define
\begin{align}
\psi_{i,0} &= h_i,\\
\psi_{i,1} &= \dot h_i+\gamma_1h_i,\\
\psi_{i,2} &=
\underline{\ddot h}_i
+(\gamma_1+\gamma_2)\dot h_i
+\gamma_1\gamma_2h_i.
\label{eq:hocbf_condition}
\end{align}
Here, $\psi_{i,2}$ uses the conservative lower bound
to account for the bounded obstacle acceleration.

At prediction stage $k$, the barrier quantities are
evaluated using $x_k$, $u_k$, and the predicted
obstacle position $\hat o_{i,k}$.
The controller incorporates these conditions
throughout the prediction horizon.
The current-step second-order
condition is enforced as
\begin{equation}
\psi_{i,2}(x_0,u_0)\geq0.
\end{equation}
The remaining barrier conditions use nonnegative
relaxation variables:
\begin{equation}
\psi_{i,r}(x_k,u_k)+s_{i,k}^{(r)}\geq0,
\qquad s_{i,k}^{(r)}\geq0,
\label{eq:relaxed_hocbf}
\end{equation}
for $r\in\{0,1,2\}$ and $k=0,\ldots,N-1$, excluding
$(k,r)=(0,2)$. Their contribution to the objective is
\begin{equation}
J_{\mathrm{slack}}
=
\rho_s
\sum_{i=1}^{n_o}
\sum_{\substack{0\leq k<N,\;0\leq r\leq2\\
(k,r)\neq(0,2)}}
\left(s_{i,k}^{(r)}\right)^2,
\end{equation}
where $\rho_s>0$ penalizes relaxation. The forward-invariance properties of HOCBFs are
established under their continuous-time assumptions,
whereas the controller is implemented at a fixed
sampling interval of $\Delta t=0.1\,\mathrm{s}$.
The current-step second-order barrier condition is
enforced as a hard constraint, while the remaining
barrier conditions are relaxed to alleviate
infeasibility.

\subsubsection{Unified Control Problem}

\label{sec:unified_control}

The resulting NMPC--HOCBF problem is
\begin{equation}
\begin{aligned}
\min_{X,U,s}\quad&
J_{\mathrm{MPC}}+J_{\mathrm{slack}}\\
\text{subject to}\\\
&x_0=x(t),\\
&x_{k+1}=F_{\Delta t}(x_k,u_k),\\
&u_k\in\mathcal{U},\qquad x_k\in\mathcal{X},\\
&\text{geometric constraints~\eqref{eq:nmpc_geometric}},\\
&\text{HOCBF constraints~\eqref{eq:relaxed_hocbf}},\\
&\psi_{i,2}(x_0,u_0)\geq0,\qquad s\geq0.
\end{aligned}
\label{eq:unified_nmpc}
\end{equation}
Here, $\mathcal{U}$ and $\mathcal{X}$ encode the input
and velocity bounds. At each control update, a nominal trajectory is
generated by rolling out the nonlinear prediction
model using the shifted previous control sequence,
or an initial tracking sequence when no previous
solution is available. A single local QP approximation
is then formed using first-order approximations of
the dynamics and nonlinear constraints. State
deviations are eliminated through the linearized
dynamics, leaving control increments and relaxation
variables as the optimization variables. Bounds on
the control increments limit departure from the
nominal sequence.
The QP is solved using OSQP \cite{stellato2020osqp}. When the primary solution
passes the controller's numerical acceptance checks,
the first control input is applied. Otherwise, the
execution procedure in Algorithm~\ref{alg:oasis}
determines the recovery or braking input.
The process is repeated at the next control update.

\begin{algorithm}[H]
\caption{OASIS navigation}
\label{alg:oasis}
\small
\begin{algorithmic}[1]
\Require Trained actor $\mu_\phi$, goal $g$, holding interval $K$
\State Reset environment and controller memory; $q\gets-1$

\While{episode is active}
    \State Build observation $z$ from robot, goal, obstacles, and $q$
    \State Generate action $\eta\gets\tanh(\mu_\phi(z))$
    \State Map $\eta$ to a subgoal and apply subgoal processing
    \State $F\gets0$; $L\gets0$

    \For{$\ell=0,\ldots,K-1$}
        \State Observe robot and obstacles; predict obstacle motion
        \State Construct and solve the NMPC--HOCBF QP
        \State $L\gets L+1$

        \If{primary solution is accepted}
            \State Extract first input $u$; update controller memory
        \Else
            \State $F\gets F+1$
            \State Attempt geometrically constrained MPC recovery
            \If{recovery solution passes validation}
                \State Extract recovery input $u$; update memory
            \Else
                \State Compute bounded braking input $u$
            \EndIf
        \EndIf

        \State Apply $u$ and record the applied input
        \State Update observations and termination conditions
        \If{episode has terminated}
            \State \textbf{break}
        \EndIf
    \EndFor

    \State Set $q\gets1$ if $F=0$, otherwise $q\gets0$
\EndWhile
\end{algorithmic}
\end{algorithm}
When the primary optimization fails, a geometrically constrained MPC recovery is attempted, followed by bounded braking if recovery is unsuccessful. These fallback actions are intended to maintain practical controller operation but do not independently provide the forward-invariance guarantees associated with the HOCBF formulation.

\section{Experiments}\label{sec:protocol}

The experiments evaluate whether the proposed method improves
navigation success. First, the hierarchical approaches are
compared with controllers that track the final goal directly.
The contribution of optimization feedback is then examined
by comparing \oasis{} with a variant that retains the same
optimization penalty but excludes feedback from the policy
observation. Additional comparisons evaluate the contribution
of the optimization penalty and the inclusion of HOCBF
constraints. Navigation success, collision avoidance,
optimization reliability, and computational cost are assessed
across five scenarios with different obstacle compositions
and uncertainty conditions.

\subsection{evaluation scenarios}

\begin{figure}
    \centering
    \includegraphics[width=0.75\linewidth]{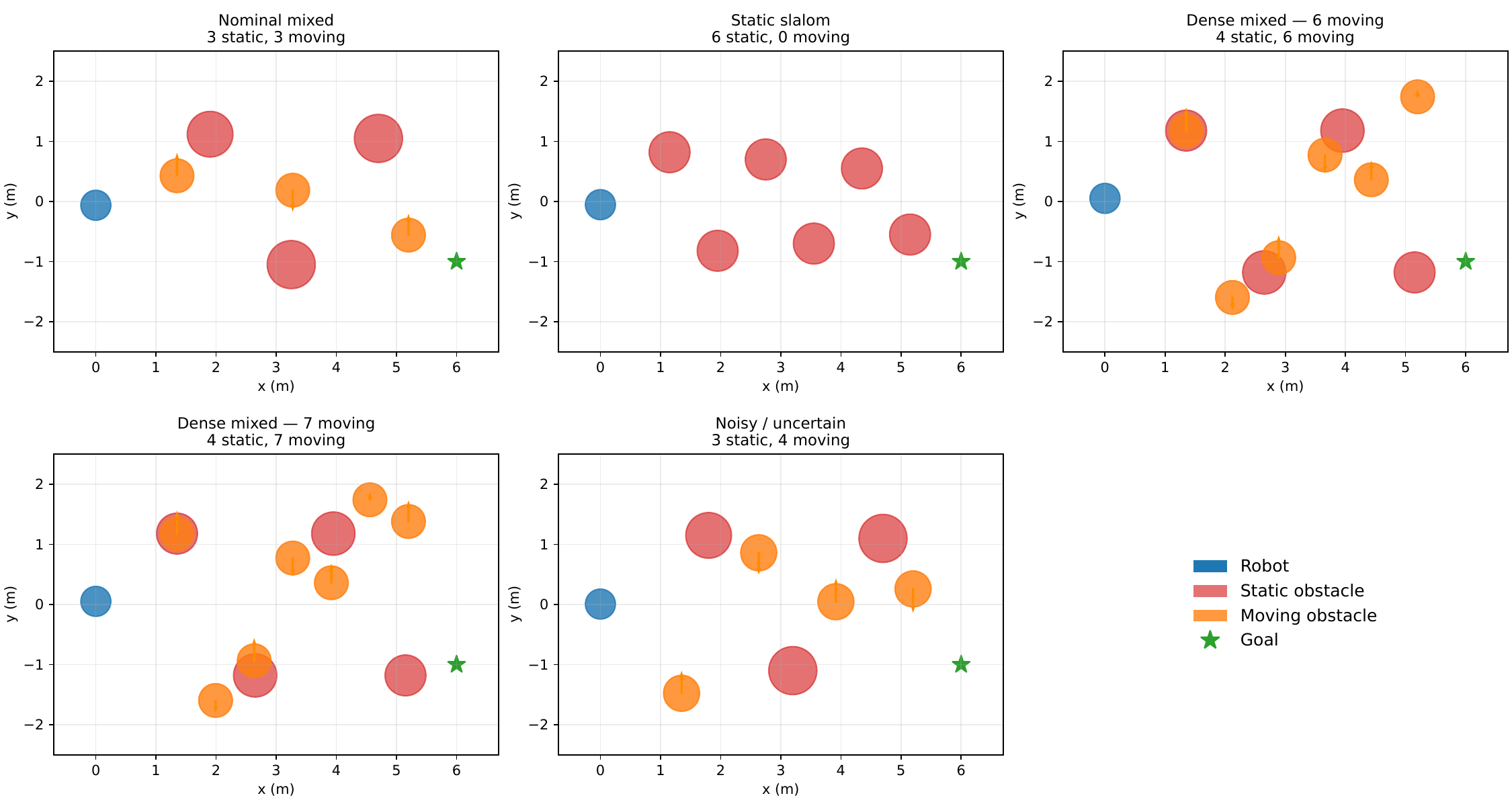}
    \caption{All scenarios for the same initial and goal positions}
    \label{fig:allscen}
\end{figure}

The nominal mixed scenario includes both static and moving
obstacles. The static slalom scenario contains static obstacles with narrow passages, testing the ability to navigate constrained
spaces. Two dense mixed scenarios increase the number of
moving obstacles, testing performance under more frequent
robot--obstacle interactions. Finally, the uncertain/noisy
scenario evaluates performance under increased sensing and
motion noise. Table~\ref{tab:scenarios} summarizes the obstacle
composition of each scenario. The environment for every scenario is demonstrated in Fig. \ref{fig:allscen}

\begin{table}[htbp]
\centering
\small
\caption{Evaluation scenarios.}
\label{tab:scenarios}
\begin{tabular}{lrr}
\toprule
Scenario & Static & Moving \\
\midrule
Nominal mixed   & 3 & 3\\
Static slalom   & 6 & 0\\
Dense-6  & 4 & 6\\
Dense-7   & 4 & 7\\
Uncertain/noisy & 3 & 4\\
\bottomrule
\end{tabular}
\end{table}

\subsection{Compared methods}

The proposed method combines SAC-generated
subgoals with an MPC controller equipped with HOCBF
constraints. Its actor receives optimization feedback,
and optimization failure is penalized in its training. A+ retains the same architecture and reward
but excludes optimization feedback from the actor's
observation. Method A additionally removes the optimization
penalty. Variants B+ and B retain the hierarchical structure but
omit HOCBF constraints from the low-level MPC controller.
B+ includes both optimization feedback and the optimization
penalty, whereas B includes neither. Finally, C and D track
the final navigation goal directly, without learned
subgoals. C uses an MPC with HOCBF constraints, whereas
D uses MPC without HOCBF constraints.
Table~\ref{tab:comparison_groups} summarizes the
configurations of all seven methods.

\begin{table}[htbp]
\centering
\small
\caption{Comparison groups.}
\label{tab:comparison_groups}
\begin{tabular}{llccc}
\toprule
Variant & Reference source & Safety Constraint & Op Feedback & Op Penalty \\
\midrule
\oasis{} & SAC subgoal & Yes & Yes & Yes \\
A+       & SAC subgoal & Yes & No  & Yes \\
A        & SAC subgoal & Yes & No & No\\
B+       & SAC subgoal & No  & Yes & Yes \\
B        & SAC subgoal & No  & No & No\\
C        & Final goal  & Yes & ---  & ---\\
D        & Final goal  & No  & --- & ---  \\
\bottomrule
\end{tabular}
\end{table}

\subsection{Training and evaluation protocol}

All learning-based methods were trained separately for
1600 episodes using identical training settings, except
for the differences defining each variant. The first
10 episodes used demonstrations generated by a
72-direction heuristic to initialize learning.
To assess sensitivity to training randomness, \oasis{},
A+, and A were each trained using three random seeds.
For each corresponding training seed, these methods
used the same scene-generation seeds to maintain comparable
training conditions. The resulting policies are evaluated
in a separate comparison to examine the consistency of
the observed performance trends. The best-performing policy
for each of these methods was then used in the main
comparison with the other algorithms.

After training, the policy parameters were fixed.
During evaluation, actions were selected deterministically
from the learned actor, without exploration noise or
heuristic blending.

Each method was evaluated over 100 episodes per scenario. Corresponding episodes
used matching seeds and shared the initial robot pose,
goal position, obstacle geometry, and dynamic-obstacle
initial states and motion parameters. Since obstacle
motion was independent of the robot, obstacle trajectories
were identical over the common simulation duration.
Sensing noise used the same distributions and initial
random-generator seeds.

At the beginning of each episode, optimization feedback was initialized to $-1$, indicating that no previous optimization outcome was available.
An episode is considered successful when the robot position satisfied
$\lVert p_t-p_{\mathrm{goal}}\rVert_2 \leq 0.30$~m,
No terminal-heading or stopping-speed
condition was imposed. A collision is detected when 
$\lVert p_t-o_{i,t}\rVert_2 \leq r_R+r_i$
for any obstacle at a sampled control step.

Episodes terminated upon successful arrival, collision,
boundary violation, or timeout. With a control interval of
$0.1$~s, the time limit was 14~s for nominal mixed and
static slalom, and 18~s for Dense-6, Dense-7, and
uncertain/noisy. Collision and boundary violation took
precedence over simultaneous goal arrival.  

Experiments were executed on a workstation equipped
with Intel Core i7-14650HX , 16 GB of RAM, and Nvidia RTX 5050.
The controller optimization problems were solved
using OSQP version 1.1.3.

All experiments used the robot model described by Riazati et al.~\cite{Riazati2025DesignAI}, with controller parameters listed in Table~\ref{tab:controller_config}.

\begin{table}[H]
\centering
\small
\caption{MPC and HOCBF configuration. HOCBF settings apply
only to HOCBF-equipped variants.}
\label{tab:controller_config}
\begin{tabular}{lll}
\toprule
Parameter & Symbol & Value \\
\midrule
Sampling interval & $\Delta t$ & $0.1$~s \\
Prediction horizon & $N$ & $15$ steps \\
Subgoal execution interval & $K$ & Up to $4$ steps \\
Robot radius & $r_R$ & $0.25$~m \\
Maximum subgoal distance & $d_{\max}$ & $1.35$~m \\
Linear speed bounds & $v$ & $[0,1.05]$~m/s \\
Angular speed bounds & $\omega$ & $[-1.8,1.8]$~rad/s \\
Linear acceleration bounds & $a$ & $[-2.8,1.6]$~m/s$^2$ \\
Angular acceleration bounds & $\alpha$ & $[-3.2,3.2]$~rad/s$^2$ \\
Geometric separation offset & $\varepsilon$ & $0.001$~m \\
Position tracking weight & --- & $5.0$ \\
Heading tracking weight & --- & $0.25$ \\
Speed tracking weight & --- & $0.35$ \\
Terminal position weight & --- & $25.0$ \\
Linear acceleration weight & --- & $0.07$ \\
Angular acceleration weight & --- & $0.05$ \\
Input-rate weight & --- & $0.1$ \\
Linear acceleration trust-region bound & --- & $3.0$~m/s$^2$ \\
Angular acceleration trust-region bound & --- & $3.2$~rad/s$^2$ \\
QP regularization & --- & $10^{-6}$ \\
Absolute and relative solver tolerances & --- & $2\times10^{-4}$ \\
Maximum solver iterations & --- & $16000$ \\
Slack acceptance tolerance & --- & $10^{-4}$ \\
\midrule
HOCBF safety margin & $m$ & $0.06$~m \\
HOCBF gains & $(\gamma_1,\gamma_2)$ & $(3.2,3.2)$~s$^{-1}$ \\
Obstacle acceleration bound & $\bar a^o$ & $0.16$~m/s$^2$ \\
Quadratic slack penalty & $\rho_s$ & $25000$ \\
\bottomrule
\end{tabular}
\end{table}

\subsection{Evaluation metrics}

Performance is measured by success, collision, and timeout
rates, with boundary violations counted as collisions.
Path length and travel time are averaged over successful
episodes only. The QP failure rate is the percentage of
controller calls in which the primary optimization fails
to return an accepted solution, regardless of subsequent
recovery. Mean controller time includes prediction, QP
construction, and solution,
but excludes policy inference, sensing, and simulation. Mean controller time is computed over all controller calls, including failed calls.

\section{Results and analysis}\label{sec:results}

Table~\ref{tab:overall_navigation_results} summarizes
performance across the named scenarios. \oasis{} achieved
the highest aggregate success rate of 77.4\%, and the lowest combined collision and
boundary-violation rate, at 19.2\%. These improvements were accompanied
by longer mean paths and travel times among successful
episodes.

\begin{table*}[h]
\centering
\caption{Navigation performance across five scenarios.
Boundary violations are counted as collisions. Path length
and travel time are averaged over successful episodes;
Controller time is averaged over control calls.}
\label{tab:overall_navigation_results}
\resizebox{\textwidth}{!}{%
\begin{tabular}{lrrrrrrr}
\toprule
Method &
Success (\%) &
Collisions (\%) &
Timeout (\%) &
Path length (m) &
Travel time (s) &
QP failure (\%) &
Controller time (ms) \\
\midrule
\oasis{}   & \textbf{77.40} & \textbf{19.20} & 3.40 & 7.84 & 9.99 & 3.94 & 36.26 \\
A+    & 70.60 & 25.00 & 4.40 & 7.59 & 9.69 & 4.51 & 36.32 \\
A     & 72.80 & 23.20 & 4.00 & 7.48 & 9.29 & 4.52 & 34.19 \\
B+    & 68.80 & 30.80 & 0.40 & 7.69 & 9.16 & \textbf{2.75} & \textbf{4.69} \\
B     & 59.60 & 39.60 & 0.80 & 8.37 & 11.09 & 3.14 & 6.84 \\
C     & 68.40 & 31.20 & 0.40 & 6.15 & 6.70 & 7.58 & 38.40 \\
D     & 59.20 & 40.80 & \textbf{0.00} & \textbf{6.03} & \textbf{6.26} & 4.98 & 5.53 \\
\bottomrule
\end{tabular}%
}
\end{table*}

\begin{table}[h]
\centering
\caption{Success rates (\%) over 100 episodes per scenario.}
\label{tab:scenario_success}
\resizebox{\columnwidth}{!}{%
\begin{tabular}{lrrrrrr}
\toprule
Method & Nominal mixed & Static slalom & Dense-6 & Dense-7 & Noisy/uncertain & Overall \\
\midrule
\oasis{} & \textbf{96} & 98 & 60 & \textbf{56} & \textbf{77} & 77.4 \\
A+       & 88 & \textbf{100} & 52 & 39 & 74 & 70.6 \\
A        & 94 & \textbf{100} & \textbf{61} & 37 & 72 & 72.8 \\
B+       & 79 & \textbf{100} & 60 & 54 & 51 & 68.8 \\
B        & 78 & 86 & 48 & 32 & 54 & 59.6 \\
C        & 82 & \textbf{100} & 56 & 43 & 61 & 68.4 \\
D        & 73 & 67 & 53 & 49 & 54 & 59.2 \\
\bottomrule
\end{tabular}%
}
\end{table}

Table~\ref{tab:scenario_success} shows how performance
varied with the navigation environment. Static slalom offered limited
separation among several methods, with the direct-goal
HOCBF controller also performing strongly. Learned subgoals
therefore provided no clear advantage in this scenario.

The dense environments revealed a different pattern.
Several configurations performed similarly in Dense-6,
whereas \oasis{} and B+ led in Dense-7. Both incorporate
optimization feedback and its associated penalty, suggesting
that this combination is useful under increased obstacle
interaction. The lower success rate of C relative to D in the dense-7
scenario indicates that additional safety constraints do not
necessarily improve task completion. One possible explanation
is that fixed-goal tracking and a limited prediction horizon
can lead the robot into situations where avoiding approaching
obstacles becomes difficult under the barrier and actuation
constraints. The higher QP failure rate of C is consistent
with this explanation

In nominal mixed and uncertain/noisy environments,
the HOCBF-equipped hierarchies outperformed their
corresponding variants without HOCBF constraints.
The stronger performance of \oasis{} relative to A+
also supports the contribution of optimization feedback
when the penalty is retained. Overall, the complete method
performed most consistently across scenarios.

\begin{table}[t]
\centering
\caption{Success rates (\%) across training seeds over five
scenarios}
\label{tab:training_seed_comparison}
\begin{tabular}{lrrr}
\toprule
Seed & \oasis{} & A+ & A \\
\midrule
42  & 75.80 & 70.60 & 71.20 \\
123 & 77.40 & 69.20 & 72.80 \\
456 & 76.20 & 69.80 & 70.80 \\
\midrule
Mean across three seeds & 76.47 & 69.87 & 71.60 \\
\bottomrule
\end{tabular}%
\end{table}

Table~\ref{tab:training_seed_comparison} examines whether
the comparison among \oasis{}, A+, and A is consistent
across training seeds. The same performance ordering
was observed for all three seeds: \oasis{} achieved
the highest success rate, followed by A and then A+.
The benefit of optimization feedback relative to the
penalty-only variant therefore persisted across the
evaluated seeds, whereas the optimization penalty alone
did not improve success relative to A. These results
support the consistency of the observed trends within
the three training seeds considered.

\begin{figure*}[t]
    \centering
    \begin{subfigure}[t]{0.32\textwidth}
        \centering
        \includegraphics[width=\linewidth]{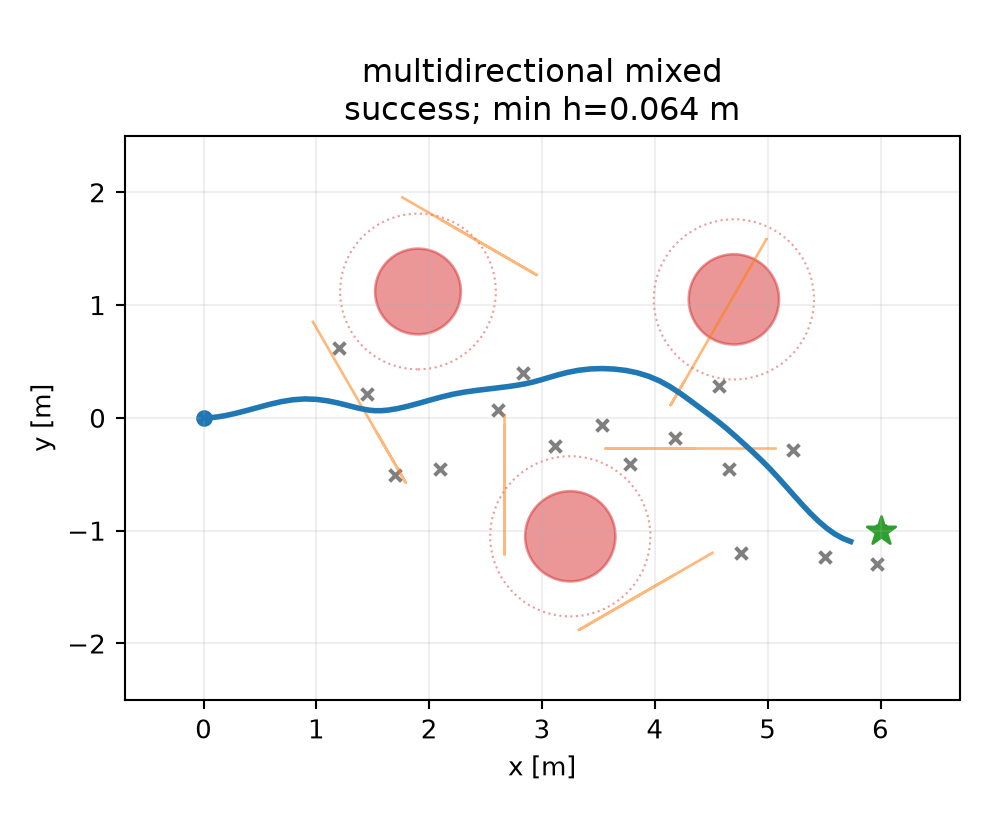}
        \caption{Nominal mixed}
        \label{fig:trajectory_nominal}
    \end{subfigure}
    \hfill
    \begin{subfigure}[t]{0.32\textwidth}
        \centering
        \includegraphics[width=\linewidth]{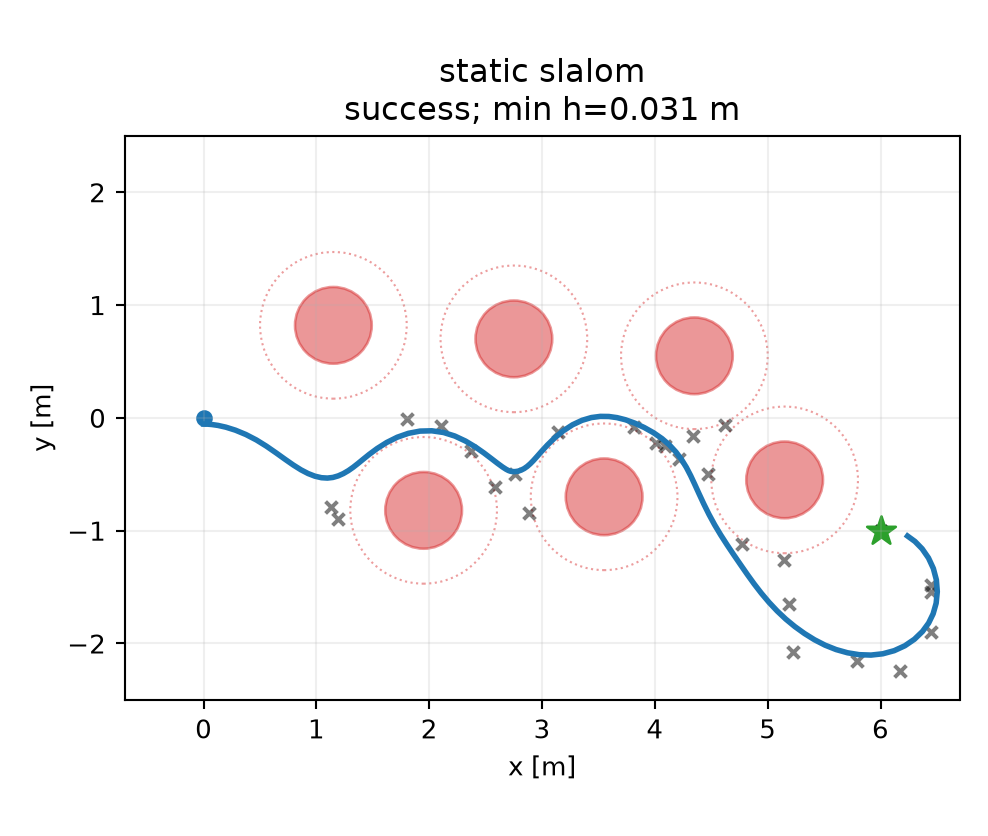}
        \caption{Static slalom}
        \label{fig:trajectory_static}
    \end{subfigure}
    \hfill
    \begin{subfigure}[t]{0.32\textwidth}
        \centering
        \includegraphics[width=\linewidth]{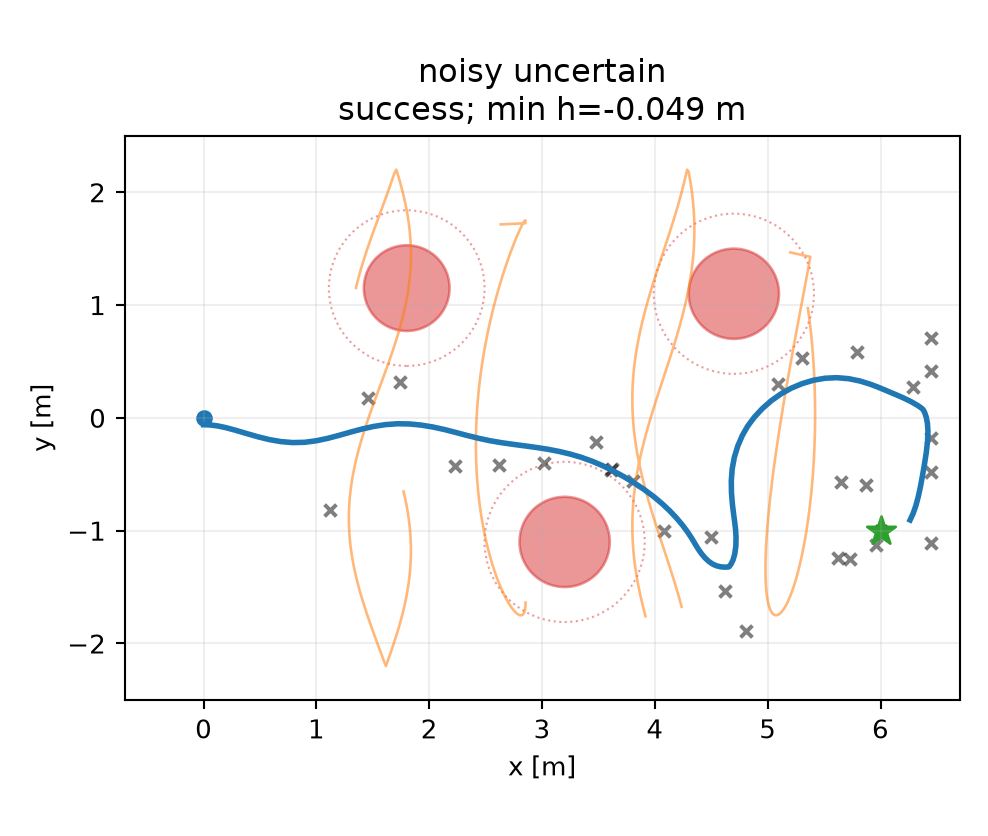}
        \caption{Uncertain/noisy}
        \label{fig:trajectory_noisy}
    \end{subfigure}

    \caption{Representative successful navigation episodes
    using \oasis{} in three evaluation scenarios.}
    \label{fig:scenario_trajectories}
\end{figure*}

Fig.~\ref{fig:scenario_trajectories} illustrates successful
navigation episodes using \oasis{} in the nominal mixed,
static slalom, and uncertain/noisy scenarios.

\section{Conclusion}

This work presented \oasis{}, a hierarchical navigation
framework combining learned subgoal generation with
NMPC and embedded HOCBF constraints. The policy receives
a summary of optimization outcomes from the preceding
subgoal interval, allowing subsequent subgoals to depend
on the controller's recent execution history.

Across the evaluated scenarios, OASIS achieved the
highest aggregate success rate. Matched comparisons
examined the contribution of the feedback observation
separately from optimization-related reward penalties.
The results indicate that previous optimization
outcomes can improve learned navigation guidance,
although the benefit varied across training seeds
and scenarios. The findings therefore support the
usefulness of this feedback.

The remaining failures highlight the challenges of
navigation under dense interactions, limited control
authority, and imperfect obstacle predictions.
Future work will investigate richer descriptions
of optimization difficulty and evaluate whether they
provide more useful guidance than the current outcome
summary. Physical robot experiments will further
assess the framework under sensing errors, execution
delays, and model mismatch.

\section*{Declaration of generative AI and AI-assisted technologies in manuscript preparation}
During preparation of this working draft, the authors used OpenAI's ChatGPT to assist with manuscript organization, LaTeX preparation assistance.  After using this tool/service, the authors reviewed and edited the content as needed and take full responsibility for the content of the published article.
\bibliographystyle{elsarticle-num}
\bibliography{references}
\end{document}